\pdfoutput=1
\documentclass[11pt]{article}

\usepackage{acl}

\usepackage{times}
\usepackage{booktabs}
\usepackage{multirow}
\usepackage{comment}
\usepackage{graphicx}
\usepackage{flushend}
\usepackage{caption}

\usepackage{latexsym}

\usepackage[T1]{fontenc}
\usepackage[utf8]{inputenc}

\usepackage{microtype}

\usepackage{graphicx}
\usepackage{booktabs}
\title{UTSA NLP at SemEval-2022 Task 4: An Exploration of Simple Ensembles of Transformers, Convolutional, and Recurrent Neural Networks}

\author{Xingmeng Zhao \and
  Anthony Rios \\
 Department of Information Systems and Cyber Security \\
  University of Texas at San Antonio \\
  San Antonio, TX\\
  \texttt{\{xingmeng.zhao, anthony.rios\}@utsa.edu} \\}

\begin{document}
\maketitle
\begin{abstract}

The act of appearing kind or helpful via the use of but having a feeling of superiority condescending and patronizing language can have have serious mental health implications to those that experience it. Thus, detecting this condescending and patronizing language online can be useful for online moderation systems. Thus, in this manuscript, we describe the system developed by Team UTSA SemEval-2022 Task 4, Detecting Patronizing and Condescending Language. Our approach explores the use of several deep learning architectures including RoBERTa, convolutions neural networks, and Bidirectional Long Short-Term Memory Networks. Furthermore, we explore simple and effective methods to create ensembles of neural network models. Overall, we experimented with several ensemble models and found that the a simple combination of five RoBERTa models achieved an F-score of .6441 on the development dataset and .5745~\footnote{Note that there was a bug in our code that resulted in different results in the official rankings.} on the final test dataset. Finally, we also performed a comprehensive error analysis to better understand the limitations of the model and provide ideas for further research.

% to detect patronizing and condescending language (PCL) in paragraphs extracted from news articles in English. Given a paragraph, systems must predict whether it contains condescending language or not (Subtask 1).
\end{abstract}

\section{Introduction}

% Problem Statement
% What is PCL? % what’s wrong with it? 
% how harmful Patronizing and Condescending Language (PCL) can be, and % ~\citep{perez2020don}

Patronizing and condescending language (PCL) generally appears as an act to hold a superior attitude, resulting in language that ``talks down'' to others. For instance, PCL may describe someone in a power position as the potential ``savior'' of a vulnerable community (e.g., ''\textit{A donation of one dollar can save a life}''), masquerading a sense of superiority as compassion. There has been recent research suggesting that PCL can have adverse effects on the mental health of individuals~\cite{giles1993patronizing,shaw2021understanding}, particularly in the context of ageism. While there has been substantial research on PCL in various contexts~\cite{huckin2002critical,komrad1983defence,giles1993patronizing,shaw2021understanding}, unfortunately, there have been few efforts to develop PCL detectors in the field of Natural Language Processing (NLP). Hence, this paper describes our team's (UTSA  NLP) contributions on the SemEval Task 4 task that introduced a new dataset for detecting PCL language. 

NLP has investigated a broad spectrum of problematic language usages, such as hate speech~\cite{vidgen2021learning}, sarcasm language~\cite{bamman2015contextualized}, fake news~\cite{hu2021compare}, and the spread of rumors and disinformation. However, PCL has only recently been explored in the NLP community~\cite{perezalmendros2020dont}. To alleviate this issue, SemEval-2022 Task 4 expanded on the work by \citet{perezalmendros2020dont}, releasing a large PCL dataset for two PCL subtasks. Subtask 1 focuses on detecting the presence of PCL in a news story. Subtask 2 asks contestants to identify the exact PCL type expressed post (if any). There are multiple technical challenges for identifying PCL. For instance, accurate models must handle imbalanced data (most news stories do not contain PCL) and complex semantic understanding to relate shallow solutions for helping vulnerable populations. For instance, \citet{perezalmendros2020dont} describe ``Shallow Solutions'' as a type of PCL, e.g., ``\textit{Raise money to combat homelessness by curling up in sleeping bags for one night}''. Nevertheless, for a model to understand this is an example of PCL, it would need to understand that ``curling up in sleeping bags for one night'' is unlikely to help the general problem of homelessness. Hence, we hypothesize that different models will learn to detect different types of PCL with varying accuracy; thus, combining multiple methods can result in better performance than a single method. 

Overall, this paper describes our system for Task 1. Specifically, we evaluate multiple combined methods to handle PCL's complex nature better than a single method. Hence, for our methodology, we trained a RoBERTa model and two traditional deep learning models (a Convolutional Neural Network and a Long Short-Term Memory Network) for comparison. In addition, we experimented with different model hyperparameters, random seeds, thresholds, and pre-trained word embeddings using the performance on the validation set to assess model variants. Finally, we evaluate multiple simple yet effective methods of combining the neural network models in an ensemble.

\section{Background}

% PCL Models 

% DataSet - Task Description
Based on the work of \citet{perezalmendros2020dont}, the SemEval Task 4 dataset contains 10,637 news stories about vulnerable people published in 20 English-speaking countries, with a novel PCL taxonomy consisting of three top-level categories (The savior, The expert, and The poet) and seven low-level PCL categories describing the different types of condescension~\cite{perez2020don}. Thet task contains two subtasks: binary classification (subtask 1) and multi-label classification (subtask 2). The binary classification for subtask 1 annotated the data with one of two categories: PCL and Not PCL. Subtask 2, the multi-label classification task, categorizes the news stories into a subset of of seven different PCL categories: unbalanced power relations, authoritative voice, shallow solution, presumption, compassion, metaphor, and the pooer, the merrier.  A complete description of each category can be found in \citet{perezalmendros2020dont}.

PCL has been studied in a wide array of contexts, from sociolinguistics to healthcare~\cite{huckin2002critical,komrad1983defence,giles1993patronizing,shaw2021understanding}. However, much of the prior work has focused on interviews and general qualitative methods. Thus, automated PCL detection models can provide social scientists with tools to understand the impact of PCL at scale. For instance, PCL models would allow linguistics to understand the implicit language actions related to condescension and aid social scientists in researching the link between condescension and other characteristics like gender or socioeconomic status because these superior attitudes and discourse of pity can routinize discrimination and make it less visible~\cite{ng2007language}. However, much of the research on harmful language in NLP has concentrated on the explicit, offensive, and apparent phenomena like false news identification, trustworthiness prediction and fact-checking, modeling offensive language, both generic and community-specific~\cite{vidgen2021learning,zampieri2019predicting,schmidt2017survey}; or how rumors spread~\cite{ma2017detect}. Recently, some work on condescending language has begun to surface. For example, based on the challenge that condescension is often undetectable from isolated discourse because it depends on discourse and social context, \citet{wang2019talkdown} introduces the task of modeling the phenomenon of condescension in direct communication from an NLP perspective and developing a dataset with annotated social media messages. Likewise, \citet{perezalmendros2020dont} also trained various baseline models to examine how existing NLP approaches perform in this task. Although they observe that recognizing PCL is achievable, it is still difficult. Hence, the work by \citet{perezalmendros2020dont} formed the basis of SemEval Task 4.% Overall, patronizing and condescending behavior helps to reinforce already existing inequities, harm vulnerable groups, and alienate persons from all walks of life. Hence, devleoping automated detection techniques can provide multiple %Unfortunately, the most vulnerable groups are the ones that are underrepresented in society. Suppose advanced NLP can correctly recognize instances of condescension or patronizing toward others. In that scenario, we may modify our language to be more inclusive and constructive and therefore more accountable for creating a more inclusive and responsible society for all.

\section{System Description}

Overall, we developed an ensemble model strategy for the PCL challenge. Specifically, we evaluated three individual methods: RoBERTa, Convolutional Neural Networks, and Long Short-Term Memory Networks. Furthermore, we experimented with various ensemble combinations. 
Approaching the task we conduct multiple experiments with a variety neural network architectures using Convolutional Neural Networks (CNN)~\citep{DBLP:journals/corr/Kim14f}, Bi-directional long short term memory (BiLSTM)~\citep{huang2015bidirectional}, and the pre-trained transformer-based model, RoBERTa~\citep{liu2019roberta}.  Each model and ensemble method is described in the section below.

\textbf{CNN.} We use the CNN model introduced by \citet{DBLP:journals/corr/Kim14f}. Intuitively, the CNN model learns to extract predictive n-grams from the text. For CNN architecture, we used filter sizes that span two, three, and four words. For the activation functions, we used ReLU~\cite{glorot2011deep}. Furthermore, we only needed two filters for each filter size~\footnote{We experimented with filter normal filter sizes from 100 to 300, but two seemed to perform just as well. We hypothesize this is because of the small number of PCL examples in the dataset.}. Between the max-pooled outputs from the convolutional layer and the the full-connected output layer, we use dropout with the probability set to 0.5 during training. The final fully-connected output layer uses a Softmax activation and outputs class probabilities for PCL or Not PCL. The model was trained with the Adam optimizer~\cite{kingma2014adam}. Furthermore, we trained the CNN models with various learning rates from 1e-4 to 1e-3 for a maximum of 35 epochs..

\textbf{BiLSTM.} While CNNs only extract informative n-grams from text, recurrent neural networks (RNNs) are able to capture long term dependencies between words. For our RNN method, we use a BiDirectional Long Short-Term Memory Network (LSTM)~\cite{hochreiter1997long}, specifically we use a variant introduced by~\newcite{graves2012supervised}. For the hyper-parameters, we did not use dropout, trained for a maximum of 35 epochs, and used variety hidden layer sizes (128, 256, and 512). The models were trained with the Adam optimizer~\cite{kingma2014adam}. Furthermore, we trained the BiLSTM models with various learning rates from 1e-4 to 1e-3.

\textbf{RoBERTa.} In our study we used a variant of BERT~\citep{devlin2018bert}, namely RoBERTa~\citep{liu2019roberta} model, which is a lighter and faster. Specifically, we use the roberta-base variant in the HuggingFace package~\cite{wolf2019huggingface}. We trained the RoBERTa model for 20 epochs with a mini-batch size set to 8 with the Adam optimizer. The learning rate was initially set to 2e-5 (other hyper-parameters same as \cite{liu2019roberta}) and the adjusted stepwise linear decay was used to modify the learning rate through training, with step sizes of two and used used.  Moreover, we used the last layer's CLS token which is passed to a final softmax layer. The model was checkpointed after each epoch, and the best version was chosen using the validation data.

\textbf{Pre-trained Word Embeddings.}
For the CNN and BiLSTM models, we compare the following pre-trained word embeddings: Word2Vec vectors trained on Google News corpus~\citep{mikolov2013distributed}, GloVe vectors trained on Wikipedia2014 and Gigaword5 corpus(GLoVe-Word) and Twitter (GLoVe-Twitter) corpora~\citep{pennington-etal-2014-glove}, and FastText vectors trained on CommonCrawl corpora~\citep{bojanowski2017enriching}. %CNN and BiLSTM models are based on different popular pre-trained embeddings,  GloVe~\citep{pennington-etal-2014-glove}, FastText~\citep{mikolov2017advances}, and SkipGram~\citep{mikolov2013distributed}. 

\textbf{Ensemble Model.}
There has been a wide array of research showing that ensembles of deep learning models have are useful for boosting  model performance~\citep{allen2020towards,peng2018extracting}. We built different ensemble models by taking an unweighted average of the probability outputs of each of the independently trained models. This includes models trained with different hyperparameters, e.g., hidden state size for the BiLSTM models, different learning rates, and different random seeds. For the CNN and BiLSTM, each model was trained on four different pre-trained word vectors described above.
The CNN and LSTM models were also trained on four separate with three random seeds, for each word embedding and learning rate combination. The RoBERTa model was trained with eleven random seeds and a number of two different stepwise learning rate schedulers using step sizes of 2 and 3. Overall, we trained a total of 46 different models for the PCL task. Next, we experimented with two methods of model averaging (ensembling). First, for Ensemble 1, we simply average the top five model instances---which resulted in different RoBERTa models trained with different random seeds and learning step settings. Second, for Ensemble 2, we experimented with taking the top five models combined with the top few CNN and BiLSTM models.

\section{Dataset, Experimental Setup, and Training Details}

For subtask 1, we use the train dataset provided by the PCL organizers. We choose the best epoch and the best hyperparameters using performance assessed in terms of F1-score on this development set. We saved the best epoch and best hyper-parameters for each model variant.
For evaluation, we use the provided test and validation datasets released by the organizers. We implemented our models on 4GPUs using Pytorch to train binary classifiers for PCL. We use Cross-Entropy Loss in all our experiments as the loss function. We ran the experiments on the server using a GPU CUDA Version: 11.4. We selected the best epoch based on the F1 score on the development set to save the best models.

% comparison
To build a basis for comparison, all models were trained using the training data provided by the task organizers and evaluated against the provided validation dataset. The best-performing models were then submitted for evaluation against the test dataset during the task evaluation period. The training method was repeated three times for CNN and LSTM and eleven times for Roberta, each with a new random seed. This is because changing the random seed used in fine-tuning RoBERTa models can provide significantly different outcomes, even if the models are similar in terms of hyper-parameters~\citep{dodge2020fine}. The best-performing hyperparameters in each model were saved for the remainder of the ensemble study.
We also found the optimal probability threshold of CNN, LSTM, and Roberta that resulted in the best F1 score on the validation dataset were .1, .35, and .35, respectively.% We also experimented with different number of models within the ensemble based on  F1 performanceto filter the best top N models, e.g., all models with f1 scores greater than .6. We perform an unweighted average of the probabilistic outputs of each independently trained model selected and validate the new models with different thresholds to balance precision and recall to achieve the best F1 score.
An optimal threshold was also chosen for the ensemble model, which was found to be .35.

\begin{table}[t]
\centering
\resizebox{.9\linewidth}{!}{
\begin{tabular}{llrrr}
\toprule
                      &                & \textbf{AVG Prec.} & \textbf{AVG Rec.} & \textbf{AVG F1} \\ \midrule
\multirow{2}{*}{\textbf{RoBERTa}} & step\_size = 2 & .5948                 & .5738              & .5826          \\
                               & step\_size = 3 & .6006                 & .5916              & .5952          \\ \midrule
\multirow{4}{*}{\textbf{CNN}}           & GoogleNews     & .2542                 & .7085              & .3733          \\
                               & FastText       & .2738                 & .5729              & .3653          \\
                               & Glove\_Word    & .2253                 & .4640               & .3031          \\
                               & Glove\_Twitter & .2070                  & .6549              & .3132          \\ \midrule
\multirow{4}{*}{\textbf{BiLSTM}}          & GoogleNews     & .3250                  & .4087              & .3609          \\
                               & FastText       & .3659                 & .4020               & .3810           \\
                               & Glove\_Word    & .3525                 & .4271              & .3831          \\
                               & Glove\_Twitter & .3745                 & .3953              & .3821          \\ \bottomrule
\end{tabular}}
\caption{Individual Models Performance}
\label{tab:indv}
\end{table}

\section{Results}

Table~\ref{tab:indv} shows the average recall, and F1 score. The scores are averaged across the different random seeds and hyperparameters used to train the models.
%The F1 score is used to select the optimal subset of the ensemble model for subtask 1.
Overall, we notise that the RoBERTa model outperforms both the CNN and BiLSTM models by more than 20\%. For the CNN model, we find that the GoogleNews word embeddings perform best. However, for the BiLSTM model, we find that the model performs similarly across all pretrained embeddings, with the Glove Word embeddings slightly outperforming others.

\begin{table}[t]
\centering
\resizebox{.9\linewidth}{!}{
\begin{tabular}{cccccc}
\toprule
               & \textbf{step\_size} & \textbf{seed} & \textbf{Prec.} & \textbf{Rec.} & \textbf{F1} \\ \midrule
\multicolumn{6}{c}{\textbf{Development Results}} \\ \midrule
\multirow{5}{*}{\textbf{RoBERTa}} & \multirow{4}{*}{3}  & 0             & .6103             & .6533          & .6311            \\
 &   & 4 & .6263 & .5980 & .6118 \\
 &   & 2 & .6029 & .6181 & .6104 \\
 &   & 7 & .6277 & .5930 & .6098 \\ \cmidrule(lr){2-6} 
 & 2 & 0 & .5980 & .6131 & .6055 \\ \midrule
\textbf{Ensemble 1}  & --- & --- & .\textbf{6215}  & .\textbf{6683} & \textbf{.6441 } \\ 
\textbf{Ensemble 2}  & --- & --- & .6093 & .6583 & .6328 \\ \midrule \midrule
\multicolumn{6}{c}{\textbf{Test Results}} \\ \midrule
\textbf{Ensemble 1}  & --- & --- & .5412 & .5804  & .5601   \\ 
\textbf{Ensemble 2}  & --- & --- &  \textbf{.5599} & \textbf{.5899 } & \textbf{.5745 } \\ \bottomrule
\end{tabular}}
\caption{Individual models in the best ensemble, and overall ensemble performance on the development and test datasets.}
\label{tab:ens}
\end{table}

In Table~\ref{tab:ens} we report the results of the two ensemble models: Ensemble 1 (only RoBERTa Models) and Ensemble 2 (Combining RoBERTa with the CNN and RNN models). On the development set, we find that that a single RoBERTa model achieves an F1 of .6311, with the next best four models achieving an F1 of around .61. The F1 of Ensemble 1 improves on the best RoBERTa models result by more than 1\%. Ensemble 2 only improves on the Best F1 by .1\%. However on the final test set, we find that the differences is not meaningful, with Ensemble 2 slightly outperforming Ensemble~1 (.56 vs. .57).

\begin{figure}[t]
    \centering
    \includegraphics[width=.9\linewidth]{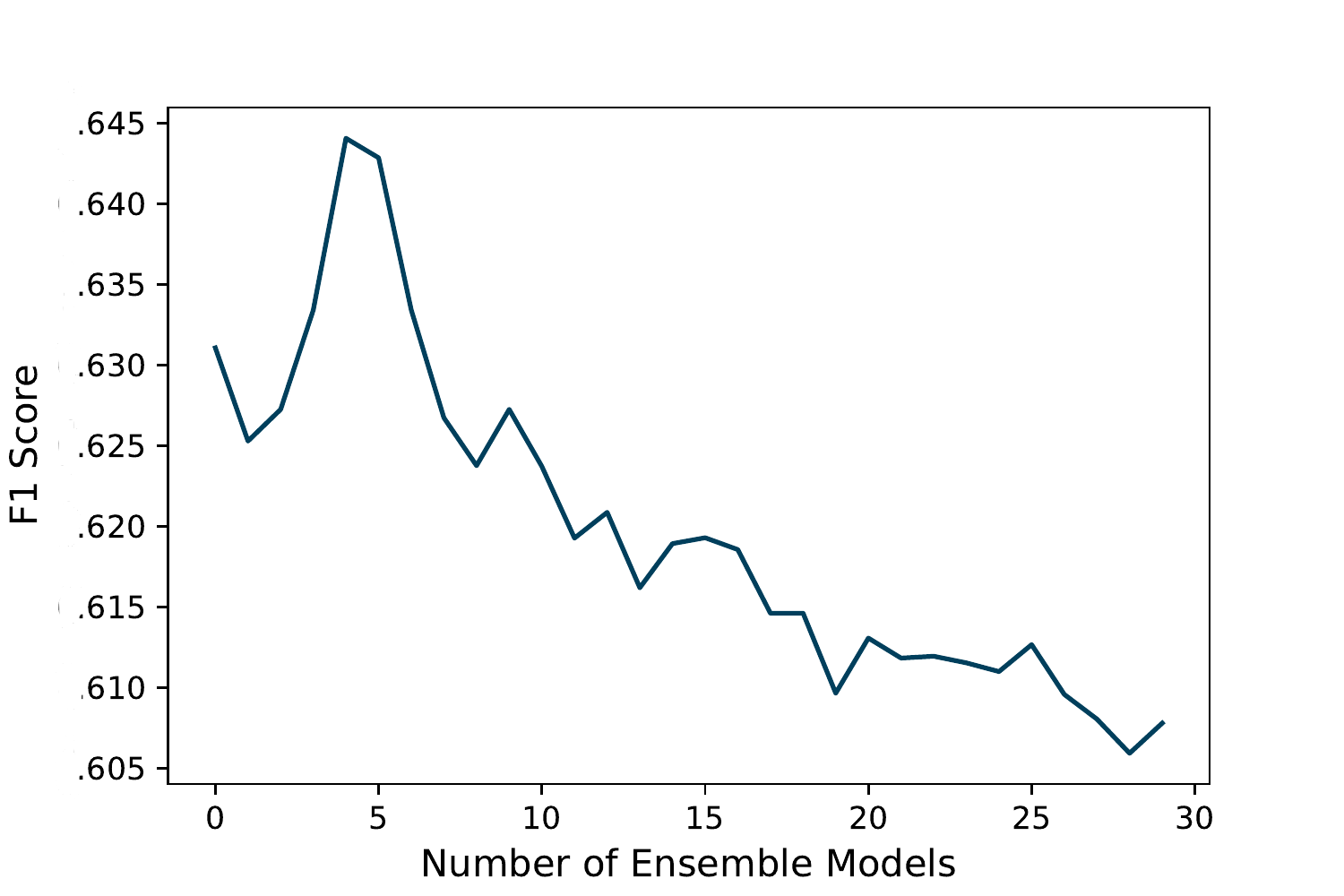}
    \caption{F1 score for different sized ensembles on the development dataset.}
    \label{fig:galaxy}
    \vspace{-1em}
\end{figure}

Next, in Figure~\ref{fig:galaxy} we report the results of averaging different number of models in our ensemble. More specifically, we evaluate averaging the best two models, best three models, and best N models, for N up to an ensemble of 30 different models. The model is chosen based on the top N performing models across all model types and random seeds. Overall, we see that initially the result of an ensemble of size 1 (i.e., only using the best RoBERTa model) has an F1 of around .63. However, that slowly increases beyond .64 at around the top five models. After that, the results slow decrease. Overall, we find that while a few models with varying performance improves the results. The more inaccurate models slowly outweigh the best performing model, thus decreasing the overall results. However, we find that the results stabilize around 0.61. Finally, in Table~\ref{ref:errors} we measure the number of False Positives and False Negatives for each of the main keywords identified in the PCL dataset (e.g., they keywords are provided by the organizers indicating a vulnerable group). The model produced 66 false negative predictions and 81 false positives predictions in total, but most of false positive errors are come from homeless,in-need, poor families, and hopeless. And the false negative error occur more frequently among the homeless, woman, immigrant, migrant and disabled topic. And there are seven class for PCL: Unbalanced power relations, Authority voice, Shallow solution, Presupposition, Compassion, Metaphor, and The poorer, the merrier. 
%Our best ensemble model leads to accuracy 92.97\% for binary PCL classification. It means that our model has erroneously labeled 70 sentences in the 1000-sentence validation set. But our F1 score is only 0.6441. To further assess the quality of our PCL detection system, we evaluated the false positive and false negative errors made on the validation set. From table 4, we found that the error analysis revealed that our model has higher error rate towards some topic than others. 

\begin{comment}
\begin{table}[t]
\centering
\begin{tabular}{@{}llcc@{}}
 &  & \multicolumn{2}{c}{Predicted} \\
 &  & PCL         & Non-PCL         \\ \cmidrule(l){3-4} 
\multicolumn{1}{c}{\multirow{2}{*}{True}} & \multicolumn{1}{c|}{PCL}     & \multicolumn{1}{c|}{133} & \multicolumn{1}{c|}{66}   \\ \cmidrule(l){3-4} 
\multicolumn{1}{c}{}                      & \multicolumn{1}{c|}{Non-PCL} & \multicolumn{1}{c|}{81}  & \multicolumn{1}{c|}{1814} \\ \cmidrule(l){3-4} 
\end{tabular}
\caption{Confusion matrix of the ensemble model with best validation F1 score, 0.6441}
\label{tab:conf}
\end{table}
\end{comment}

\subsection{False Positives and False Negatives}

In addition to an exploration of the observation results, we perform an error analysis by manually comparing the true labels and predictions of Ensemble 1. First, for False Positives, we analyze an example related to ``hopelessness/homelessness'':
\begin{center}
\noindent\fbox{%
    \parbox{.93\linewidth}{%
        FP Example:``The City Without Drugs organisation is still active , as is their YouTube channel . It features hundreds of videos of drug addicts being dragged half-conscious through the street , their faces not blurred , or confessing their alleged worthlessness , their hopelessness , their shame.''
    }
}
\end{center}
This paragraphs is predicted as PCL, but the ground-truth is Not PCL. This example indicates that the Ensemble incorrectly identifies sentences as PCL when they contain many PCL-related words that may be related to PCL-like text (e.g., related to hopelessness), even when the text is not directly indicating a feeling of superiority.  Another false positive example is from the ``homeless'' topics:
\begin{center}
\noindent\fbox{%
    \parbox{.93\linewidth}{%
        FP Example:``Viral photo helping fund homeless kid , his dog.''
    }
}
\end{center}
We can see that the entity of this sentence is a single individual. This paragraph is recognized as PCL by the system, maybe because the PCL system believes it contains the shallow solution (i.e., viral photo). However, it neglected the fact that financing a specific homeless child may be realistic, i.e. it may not be a shallow solution for a single person.  \citet{perezalmendros2020dont} also mention that shallow solutions are also often overlooked by RoBERTa, where recognizing shallow solutions in the text requires external knowledge of the situation and the needs of those affected. Thus, a large number of false positive results are generated by misidentifying the entities and the relationship between patronizing and condescending language. Next, we look at a False Negative:
\begin{center}
\noindent\fbox{%
    \parbox{.93\linewidth}{%
        FN Example: ``Charity plans to forgo parking so homeless can have gym and medical centre.''
    }
}
\end{center}
Here we find another issue with shallow solutions. Specifically, the model is not able to associate the proposed procedure as not being a method of really addressing homelessness. Specifically, the PCL system unaware that ``forgoing parking'' is not a complete solution to help homeless people, which is a simple and superficial philanthropic effort that is unlikely to make a significant difference on vulnerable communities. The second example of a false negative concerns presuppositions. People need to decide whether the assumption made is reasonable or not for this type of PCL. We found for the political topic, like immigrant and migrant topics, have a lot assumptions. For example, in the following situation, the author assumes Filipino families are poor and need assistance based on stereotypes. 
\begin{center}
\noindent\fbox{%
    \parbox{.93\linewidth}{%
        FN Example:``But if the Supreme Court gives a favorable decision for the president , his immigration program would immediately take effect , changing the lives of eligible Filipino families and other immigrants.''
    }
}
\end{center}
This again suggests that the model is incapable of understanding complex relatinships between vulnerable communities and ideas.
A future interesting research avenue would explore methods incorporating relevant knowledge bases, similar to recent work on common sense generation~\cite{xing2021km}, into transformer models to address these errors.

\begin{table}[t]
\centering
\resizebox{0.9\linewidth}{!}{
\begin{tabular}{lrrrr}
\toprule
 & \textbf{FP Count} & \textbf{FN Count} & \textbf{Total PCL} & \textbf{Total} \\ \midrule
\textbf{homeless} & 15 & 9 & 29 & 212 \\
\textbf{poor-families} & 13 & 17 & 38 & 190 \\
\textbf{women} & 3 & 8 & 14 & 233 \\
\textbf{in-need} & 15 & 2 & 33 & 226 \\
\textbf{immigrant} & 0 & 4 & 7 & 218 \\
\textbf{hopeless} & 15 & 9 & 26 & 217 \\
\textbf{vulnerable} & 6 & 4 & 20 & 209 \\
\textbf{migrant} & 2 & 3 & 5 & 207 \\
\textbf{disabled} & 4 & 7 & 14 & 194 \\
\textbf{refugee} & 8 & 3 & 13 & 188 \\ \bottomrule
\end{tabular}}
\caption{Errors}
\vspace{-1em}
\label{ref:errors}
\end{table}

\section{Conclusion}

In this paper, we have presented our submission for the PCL detection system submitted to the SemEval-2022 Task 4. Our team are focus on the subtask1 to identify whether the paragraphs contain the PCL or not. We proposed several ensemble model that leverages pretrained word vector and three different deep learning architectures. In future efforts, we plan to further improve our model by incorporating structured knowledge bases.

\section*{Acknowledgements}

This material is based upon work supported by the National Science Foundation under Grant No. 1947697.

\bibliography{custom,anthology}

\begin{thebibliography}{30}
\expandafter\ifx\csname natexlab\endcsname\relax\def\natexlab#1{#1}\fi

\bibitem[{Allen-Zhu and Li(2020)}]{allen2020towards}
Zeyuan Allen-Zhu and Yuanzhi Li. 2020.
\newblock Towards understanding ensemble, knowledge distillation and
  self-distillation in deep learning.
\newblock \emph{arXiv preprint arXiv:2012.09816}.

\bibitem[{Bamman and Smith(2015)}]{bamman2015contextualized}
David Bamman and Noah Smith. 2015.
\newblock Contextualized sarcasm detection on twitter.
\newblock In \emph{Proceedings of the International AAAI Conference on Web and
  Social Media}, volume~9, pages 574--577.

\bibitem[{Bojanowski et~al.(2017)Bojanowski, Grave, Joulin, and
  Mikolov}]{bojanowski2017enriching}
Piotr Bojanowski, Edouard Grave, Armand Joulin, and Tomas Mikolov. 2017.
\newblock Enriching word vectors with subword information.
\newblock \emph{Transactions of the Association for Computational Linguistics},
  5:135--146.

\bibitem[{Devlin et~al.(2018)Devlin, Chang, Lee, and
  Toutanova}]{devlin2018bert}
Jacob Devlin, Ming-Wei Chang, Kenton Lee, and Kristina Toutanova. 2018.
\newblock Bert: Pre-training of deep bidirectional transformers for language
  understanding.
\newblock \emph{arXiv preprint arXiv:1810.04805}.

\bibitem[{Dodge et~al.(2020)Dodge, Ilharco, Schwartz, Farhadi, Hajishirzi, and
  Smith}]{dodge2020fine}
Jesse Dodge, Gabriel Ilharco, Roy Schwartz, Ali Farhadi, Hannaneh Hajishirzi,
  and Noah Smith. 2020.
\newblock Fine-tuning pretrained language models: Weight initializations, data
  orders, and early stopping.
\newblock \emph{arXiv preprint arXiv:2002.06305}.

\bibitem[{Giles et~al.(1993)Giles, Fox, and Smith}]{giles1993patronizing}
Howard Giles, Susan Fox, and Elisa Smith. 1993.
\newblock Patronizing the elderly: Intergenerational evaluations.
\newblock \emph{Research on Language and Social Interaction}, 26(2):129--149.

\bibitem[{Glorot et~al.(2011)Glorot, Bordes, and Bengio}]{glorot2011deep}
Xavier Glorot, Antoine Bordes, and Yoshua Bengio. 2011.
\newblock Deep sparse rectifier neural networks.
\newblock In \emph{Proceedings of the fourteenth international conference on
  artificial intelligence and statistics}, pages 315--323. JMLR Workshop and
  Conference Proceedings.

\bibitem[{Graves(2012)}]{graves2012supervised}
Alex Graves. 2012.
\newblock \emph{Supervised sequence labelling with recurrent neural networks},
  volume 385.
\newblock Springer.

\bibitem[{Hochreiter and Schmidhuber(1997)}]{hochreiter1997long}
Sepp Hochreiter and J{\"u}rgen Schmidhuber. 1997.
\newblock Long short-term memory.
\newblock \emph{Neural computation}, 9(8):1735--1780.

\bibitem[{Hu et~al.(2021)Hu, Yang, Zhang, Zhong, Tang, Shi, Duan, and
  Zhou}]{hu2021compare}
Linmei Hu, Tianchi Yang, Luhao Zhang, Wanjun Zhong, Duyu Tang, Chuan Shi, Nan
  Duan, and Ming Zhou. 2021.
\newblock Compare to the knowledge: Graph neural fake news detection with
  external knowledge.
\newblock In \emph{Proceedings of the 59th Annual Meeting of the Association
  for Computational Linguistics and the 11th International Joint Conference on
  Natural Language Processing (Volume 1: Long Papers)}, pages 754--763.

\bibitem[{Huang et~al.(2015)Huang, Xu, and Yu}]{huang2015bidirectional}
Zhiheng Huang, Wei Xu, and Kai Yu. 2015.
\newblock Bidirectional lstm-crf models for sequence tagging.
\newblock \emph{arXiv preprint arXiv:1508.01991}.

\bibitem[{Huckin(2002)}]{huckin2002critical}
Thomas Huckin. 2002.
\newblock Critical discourse analysis and the discourse of condescension.
\newblock \emph{Discourse studies in composition}, 155:176.

\bibitem[{Kim(2014)}]{DBLP:journals/corr/Kim14f}
Yoon Kim. 2014.
\newblock \href {http://arxiv.org/abs/1408.5882} {Convolutional neural networks
  for sentence classification}.
\newblock \emph{CoRR}, abs/1408.5882.

\bibitem[{Kingma and Ba(2014)}]{kingma2014adam}
Diederik~P Kingma and Jimmy Ba. 2014.
\newblock Adam: A method for stochastic optimization.
\newblock \emph{arXiv preprint arXiv:1412.6980}.

\bibitem[{Komrad(1983)}]{komrad1983defence}
Mark~S Komrad. 1983.
\newblock A defence of medical paternalism: maximising patients' autonomy.
\newblock \emph{Journal of medical ethics}, 9(1):38--44.

\bibitem[{Liu et~al.(2019)Liu, Ott, Goyal, Du, Joshi, Chen, Levy, Lewis,
  Zettlemoyer, and Stoyanov}]{liu2019roberta}
Yinhan Liu, Myle Ott, Naman Goyal, Jingfei Du, Mandar Joshi, Danqi Chen, Omer
  Levy, Mike Lewis, Luke Zettlemoyer, and Veselin Stoyanov. 2019.
\newblock Roberta: A robustly optimized bert pretraining approach.
\newblock \emph{arXiv preprint arXiv:1907.11692}.

\bibitem[{Ma et~al.(2017)Ma, Gao, and Wong}]{ma2017detect}
Jing Ma, Wei Gao, and Kam-Fai Wong. 2017.
\newblock Detect rumors in microblog posts using propagation structure via
  kernel learning.
\newblock In \emph{Proceedings of the 55th Annual Meeting of the Association
  for Computational Linguistics (Volume 1: Long Papers)}, pages 708--717.

\bibitem[{Mikolov et~al.(2013)Mikolov, Sutskever, Chen, Corrado, and
  Dean}]{mikolov2013distributed}
Tomas Mikolov, Ilya Sutskever, Kai Chen, Greg~S Corrado, and Jeff Dean. 2013.
\newblock Distributed representations of words and phrases and their
  compositionality.
\newblock \emph{Advances in neural information processing systems}, 26.

\bibitem[{Ng(2007)}]{ng2007language}
Sik~Hung Ng. 2007.
\newblock Language-based discrimination: Blatant and subtle forms.
\newblock \emph{Journal of Language and Social Psychology}, 26(2):106--122.

\bibitem[{Peng et~al.(2018)Peng, Rios, Kavuluru, and Lu}]{peng2018extracting}
Yifan Peng, Anthony Rios, Ramakanth Kavuluru, and Zhiyong Lu. 2018.
\newblock Extracting chemical--protein relations with ensembles of svm and deep
  learning models.
\newblock \emph{Database}, 2018.

\bibitem[{Pennington et~al.(2014)Pennington, Socher, and
  Manning}]{pennington-etal-2014-glove}
Jeffrey Pennington, Richard Socher, and Christopher Manning. 2014.
\newblock \href {https://doi.org/10.3115/v1/D14-1162} {{G}lo{V}e: Global
  vectors for word representation}.
\newblock In \emph{Proceedings of the 2014 Conference on Empirical Methods in
  Natural Language Processing ({EMNLP})}, pages 1532--1543, Doha, Qatar.
  Association for Computational Linguistics.

\bibitem[{P{\'e}rez-Almendros et~al.(2020{\natexlab{a}})P{\'e}rez-Almendros,
  Espinosa-Anke, and Schockaert}]{perez2020don}
Carla P{\'e}rez-Almendros, Luis Espinosa-Anke, and Steven Schockaert.
  2020{\natexlab{a}}.
\newblock Don't patronize me! an annotated dataset with patronizing and
  condescending language towards vulnerable communities.
\newblock \emph{arXiv preprint arXiv:2011.08320}.

\bibitem[{P{\'e}rez-Almendros et~al.(2020{\natexlab{b}})P{\'e}rez-Almendros,
  Espinosa-Anke, and Schockaert}]{perezalmendros2020dont}
Carla P{\'e}rez-Almendros, Luis Espinosa-Anke, and Steven Schockaert.
  2020{\natexlab{b}}.
\newblock {Don’t Patronize Me! An Annotated Dataset with Patronizing and
  Condescending Language towards Vulnerable Communities}.
\newblock In \emph{Proceedings of the 28th International Conference on
  Computational Linguistics}, pages 5891--5902.

\bibitem[{Schmidt and Wiegand(2017)}]{schmidt2017survey}
Anna Schmidt and Michael Wiegand. 2017.
\newblock A survey on hate speech detection using natural language processing.
\newblock \emph{SocialNLP 2017}, page~1.

\bibitem[{Shaw and Gordon(2021)}]{shaw2021understanding}
Clarissa~A Shaw and Jean~K Gordon. 2021.
\newblock Understanding elderspeak: An evolutionary concept analysis.
\newblock \emph{Innovation in aging}, 5(3):igab023.

\bibitem[{Vidgen et~al.(2021)Vidgen, Thrush, Waseem, and
  Kiela}]{vidgen2021learning}
Bertie Vidgen, Tristan Thrush, Zeerak Waseem, and Douwe Kiela. 2021.
\newblock Learning from the worst: Dynamically generated datasets to improve
  online hate detection.
\newblock In \emph{Proceedings of the 59th Annual Meeting of the Association
  for Computational Linguistics and the 11th International Joint Conference on
  Natural Language Processing (Volume 1: Long Papers)}, pages 1667--1682.

\bibitem[{Wang and Potts(2019)}]{wang2019talkdown}
Zijian Wang and Christopher Potts. 2019.
\newblock Talkdown: A corpus for condescension detection in context.
\newblock \emph{arXiv preprint arXiv:1909.11272}.

\bibitem[{Wolf et~al.(2019)Wolf, Debut, Sanh, Chaumond, Delangue, Moi, Cistac,
  Rault, Louf, Funtowicz et~al.}]{wolf2019huggingface}
Thomas Wolf, Lysandre Debut, Victor Sanh, Julien Chaumond, Clement Delangue,
  Anthony Moi, Pierric Cistac, Tim Rault, R{\'e}mi Louf, Morgan Funtowicz,
  et~al. 2019.
\newblock Huggingface's transformers: State-of-the-art natural language
  processing.
\newblock \emph{arXiv preprint arXiv:1910.03771}.

\bibitem[{Xing et~al.(2021)Xing, Shi, Meng, Lakemeyer, Ma, and
  Wattenhofer}]{xing2021km}
Yiran Xing, Zai Shi, Zhao Meng, Gerhard Lakemeyer, Yunpu Ma, and Roger
  Wattenhofer. 2021.
\newblock Km-bart: Knowledge enhanced multimodal bart for visual commonsense
  generation.
\newblock In \emph{Proceedings of the 59th Annual Meeting of the Association
  for Computational Linguistics and the 11th International Joint Conference on
  Natural Language Processing (Volume 1: Long Papers)}, pages 525--535.

\bibitem[{Zampieri et~al.(2019)Zampieri, Malmasi, Nakov, Rosenthal, Farra, and
  Kumar}]{zampieri2019predicting}
Marcos Zampieri, Shervin Malmasi, Preslav Nakov, Sara Rosenthal, Noura Farra,
  and Ritesh Kumar. 2019.
\newblock Predicting the type and target of offensive posts in social media.
\newblock In \emph{Proceedings of the 2019 Conference of the North American
  Chapter of the Association for Computational Linguistics: Human Language
  Technologies, Volume 1 (Long and Short Papers)}, pages 1415--1420.

\end{thebibliography}
\bibliographystyle{acl_natbib}

%\appendix

%\section{Example Appendix}
%\label{sec:appendix}

%This is an appendix.

\end{document}